\pdfoutput=1
\documentclass[11pt]{article}
\usepackage[preprint]{acl}
\usepackage{times}
\usepackage{latexsym}
\usepackage[T1]{fontenc}
\usepackage[utf8]{inputenc}
\usepackage{microtype}
\usepackage{inconsolata}
\usepackage{graphicx}
\usepackage{booktabs}
\usepackage{amsmath,amssymb}
\usepackage{xcolor}

\newcommand{\baseAcc}{0.573}
\newcommand{\baseConf}{0.963}
\newcommand{\baseAUROC}{0.711}
\newcommand{\baseECEraw}{0.391}
\newcommand{\baseECEcal}{0.074}
\newcommand{\baseTzero}{10.5}
\newcommand{\nItems}{600}
\newcommand{\baseSaturated}{283}

\newcommand{\baseAP}{0.782}

\newcommand{\baseTopTwenty}{0.867}
\newcommand{\ceAPdropMin}{0.029}
\newcommand{\ceAPdropMax}{0.050}

\newcommand{\ceAUROC}{0.704}

\newcommand{\ceDrift}{+0.049}
\newcommand{\ceDriftsd}{0.006}
\newcommand{\ceAcc}{0.567}
\newcommand{\ceKept}{0.795}
\newcommand{\ceTracked}{0.090}
\newcommand{\ceStale}{0.090}
\newcommand{\ceOracle}{0.089}
\newcommand{\ceRawdECE}{+0.015}
\newcommand{\ceAUROCtracked}{0.699}
\newcommand{\ceAP}{0.741}

\newcommand{\ceTopTwenty}{0.800}

\newcommand{\halfAUROC}{0.721}

\newcommand{\halfDrift}{$-$0.003}

\newcommand{\halfKept}{0.881}

\newcommand{\oneepAUROC}{0.727}

\newcommand{\oneepDrift}{$-$0.017}
\newcommand{\oneepDriftsd}{0.004}

\newcommand{\oneepKept}{0.944}

\newcommand{\oneepAP}{0.781}

\newcommand{\oneeppAP}{0.010}

\newcommand{\tceAUROC}{0.714}

\newcommand{\tceDrift}{$-$0.019}

\newcommand{\tceAP}{0.767}

\newcommand{\tcepAP}{0.039}

\newcommand{\entAUROC}{0.715}

\newcommand{\entDrift}{+0.000}

\newcommand{\focalAUROC}{0.742}

\newcommand{\focalDrift}{$-$0.132}

\newcommand{\focalTracked}{0.094}
\newcommand{\focalStale}{0.147}
\newcommand{\focalOracle}{0.091}

\newcommand{\focalAUROCtracked}{0.750}
\newcommand{\focalAP}{0.811}

\newcommand{\focaldMargin}{+0.035}
\newcommand{\focalpMargin}{0.013}
\newcommand{\focaldAUROC}{+0.038}
\newcommand{\focalpAUROC}{0.009}

\newcommand{\ktwoAUROC}{0.742}

\newcommand{\ktwoDrift}{$-$0.069}

\newcommand{\ktwoAP}{0.798}

\newcommand{\ktwodMargin}{+0.037}
\newcommand{\ktwopMargin}{0.013}

\newcommand{\ktwopAUROC}{0.011}

\newcommand{\mixAUROC}{0.720}

\newcommand{\mixDrift}{$-$0.190}
\newcommand{\mixDriftsd}{0.001}
\newcommand{\mixAcc}{0.569}

\newcommand{\mixTracked}{0.074}

\newcommand{\mixRawdECE}{$-$0.072}

\newcommand{\mixAP}{0.784}

\newcommand{\mixpAUROC}{0.261}

\newcommand{\holdMaxpDrift}{0.024}
\newcommand{\holdMinpAUROC}{0.176}
\newcommand{\baseOverconf}{0.390}
\newcommand{\focalVsHalfdAUROC}{+0.022}
\newcommand{\focalVsHalfp}{0.127}
\newcommand{\ancKept}{0.856}

\newcommand{\eoneancDrift}{$-$0.073}

\newcommand{\kfourancDrift}{$-$0.129}

\newcommand{\lsTracked}{0.102}
\newcommand{\lsOracle}{0.081}

\newcommand{\svtopKept}{0.830}
\newcommand{\svtopDrift}{+0.046}

\newcommand{\trackMaxGap}{0.010}

\newcommand{\focaldAUROCtracked}{+0.051}
\newcommand{\focalpAUROCtracked}{0.001}
\newcommand{\eoneancEntropyFirst}{0.292}
\newcommand{\eoneancEntropyLast}{0.296}
\newcommand{\ceEntropyFirst}{0.293}
\newcommand{\ceEntropyLast}{0.117}
\newcommand{\ceEntropyLost}{60}
\newcommand{\probeKeepGzero}{81}
\newcommand{\probeKeepGfive}{64}
\newcommand{\probeGreedyGzero}{73}
\newcommand{\probeGreedyGfive}{90}
\newcommand{\eightrealStale}{0.247}
\newcommand{\eightrealTracked}{0.057}
\newcommand{\eightrealOracle}{0.059}

\newcommand{\gen}[1]{g_{#1}}

\title{Temperature, Not Knowledge: What Recursive Self-Training\\and Its Remedies Do to a Language Model's Confidence}

\author{Dipto Sumit, Syeda Fabiayee Jahan, Farig Sadeque \\
  BRAC University, Dhaka, Bangladesh \\
  \texttt{dipto.sumit@g.bracu.ac.bd}}

\begin{document}
\maketitle

\begin{abstract}
Recursive self-training can collapse language models, and many proposed remedies target overconfidence, yet their effects on confidence are rarely measured. We study a five-round self-training loop on Qwen3-1.7B, using three seeds per compared method, and distinguish confidence \emph{temperature} (drift from a calibrated starting point) from \emph{discrimination} (ranking correct answers above incorrect ones). Accuracy and overall discrimination remain nearly unchanged, but confidence sharpens and the most confident answers become less reliable. Matching predictive entropy to the calibrated starting model each round recalibrates nearly as well as label-based temperature fitting, at both 1.7B and 8B. Sharpening is not inherent to self-training: exact samples from a model's distribution have zero expected gradient. Instead, most drift comes from fitting each generated answer twice, with top-$p$ truncation contributing the rest; removing both keeps entropy on generated outputs flat. Remedies affect the two properties differently: truncated cross-entropy, an entropy bonus, and single-pass training preserve temperature without improving discrimination, while focal loss and multi-sample targets improve discrimination but cause underconfidence. Reference-data mixing also overshoots. Pairing a ranking-improving loss with label-free recalibration achieves both.
\end{abstract}

\section{Introduction}

Language models now routinely learn from text that models wrote. Some pipelines do this on purpose, fine-tuning a model on its own answers to improve it or to avoid the cost of labelling \citep{huang2023selfimprove}, and any model trained on recent web text does it by accident. When the recursion runs for several rounds, the model can collapse: the learned distribution narrows, rare content disappears, and eventually task performance falls \citep{shumailov2024,alemohammad2024mad,dohmatob2024tails}.

Much of the work on preventing collapse blames one mechanism. The model becomes too sure of its own outputs, and each round of training on them makes it surer. ForTIFAI's confidence-aware losses skip or down-weight the tokens a model already predicts with high confidence, on the grounds that overconfidence in self-generated data causes feedback loops that degrade performance \citep{shabgahi2025fortifai}. Self-verification keeps only the synthetic data the model is confident about, and provably prevents collapse in the setting its authors analyse \citep{fu2025selfverification}. Yet these methods are judged by knowledge retention, benchmark accuracy, perplexity and divergence from the original distribution. None of these measures looks at confidence, the quantity the methods were designed around.

Confidence deserves a direct look because it is what the users of a deployed model act on. A self-trained model that answers as accurately as before but is more certain of its mistakes has become worse, in a way that no accuracy benchmark will reveal. In this paper we ask what recursive self-training, and the remedies proposed for it, do to a model's confidence.\footnote{Code and experiment scripts: \url{https://anonymous.4open.science/r/quant_forget-B5D4/}}

We study the question in a deliberately simple loop. A model answers a fixed pool of 3{,}000 questions and is fine-tuned on its own answers, and the result answers the same questions in the next round. We compare the models from all six rounds on the same \nItems\ multiple-choice items. A single calibration number turns out to be a poor guide here, because it mixes two properties of confidence that self-training affects differently. The first is \emph{temperature}, how confident the model is overall, which can be corrected after training by rescaling its logits. The second is \emph{discrimination}, whether higher confidence goes with correct answers, which rescaling cannot in general repair. We measure discrimination by AUROC on exact log-probabilities. We measure temperature as drift away from a calibrated starting point, because our base model is so overconfident that its raw calibration error improves under any intervention that softens it, whether or not its confidence becomes more trustworthy.

\begin{figure}[t]
\centering
\includegraphics[width=\columnwidth]{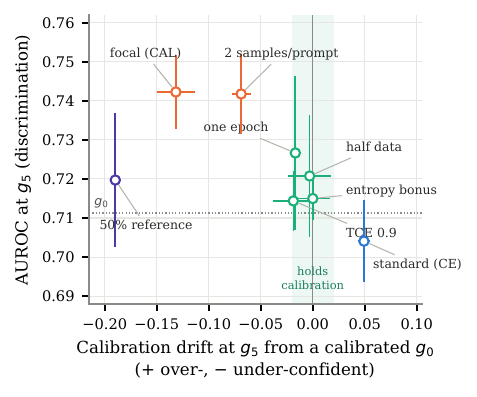}
\caption{Each replicated protocol on the two properties of confidence (mean $\pm$ sd over three seeds). Protocols in green keep a calibrated model calibrated (shaded band, $|$drift$|\le 0.02$); those in orange improve discrimination; violet is 50\% reference data. The dotted line marks the starting model $\gen{0}$. No protocol reaches the upper centre, where both properties would be good.}
\label{fig:axes}
\end{figure}

Our first finding is that self-training changes the model's temperature far more than its knowledge. Accuracy and AUROC stay close to where they started, while the model becomes steadily more confident than it should be. The ranking does suffer at the top, where the answers the model is most sure of become less reliable. Because most of the change is a shift in temperature, most of it can be undone without labels. Choosing each round's temperature so that the model's predictive entropy matches that of the calibrated starting model (adapting \citealp{luo2025pretrained}) recalibrates almost as well as refitting the temperature on labels, at 1.7B and at 8B (\S\ref{sec:temp}).

Our second finding concerns where the sharpening comes from. When a model is trained on exact samples from its own distribution, the expected gradient is zero, so self-training does not have to sharpen anything. The sharpening we observe comes from two common design choices that break this condition. Each self-generated answer is fitted twice, which accounts for most of the effect, and the answers are sampled with top-$p$ truncation, which accounts for the rest. With a single pass over exact samples, the model's entropy on its own outputs stays flat for five rounds (\S\ref{sec:mech}).

Third, each remedy we test moves one property and not the other (Figure~\ref{fig:axes}). Truncated cross-entropy, an entropy bonus and single-pass training keep a calibrated model calibrated, but they do not improve its ranking. A focal loss from the same confidence-aware family and training on two samples per prompt improve AUROC by \focaldAUROC, but they push the model into underconfidence. Mixing in reference data, the standard defence against collapse, overshoots as well, and selecting training data by the model's own confidence makes overconfidence worse in a single-seed test (\S\ref{sec:remedies}). The overshoot of the ranking-improving losses is mostly a temperature shift, so pairing such a loss with label-free recalibration obtains both properties (\S\ref{sec:both}).

\section{Related Work}

\paragraph{Model collapse and its remedies.}
Training recursively on generated data narrows the learned distribution and loses its tails \citep{shumailov2024,alemohammad2024mad,dohmatob2024tails}. Accumulating real data alongside the synthetic data, rather than replacing it, prevents collapse \citep{gerstgrasser2024,bertrand2024stability}, and so can verifying or correcting the synthetic data before training on it \citep{feng2024verification,gillman2024selfcorrecting}. \citet{yi2025escaping} show that an external verifier brings early gains that level off or reverse unless the verifier is perfectly reliable. A second line of work targets the model's confidence directly. ForTIFAI's truncated cross-entropy (TCE) ignores tokens the model predicts above a confidence threshold and tolerates 2.3 times more synthetic data before collapse; the same paper studies a family of confidence-aware losses that includes a focal loss \citep{shabgahi2025fortifai}. \citet{fu2025selfverification} prove that filtering synthetic data by the model's internal confidence prevents collapse, while \citet{qiao2026selection} show that selection can instead accelerate collapse when the selected data are biased. All of these studies measure collapse. We measure what the same remedies do to confidence.

\paragraph{Confidence under fine-tuning.}
Calibration and accuracy are distinct properties \citep{guo2017}, and instruction tuning and alignment change how well an LLM's confidence matches its accuracy \citep{kadavath2022know,tian2023justask,huang-etal-2026-investigating}. In continual fine-tuning, uncertainty coverage degrades several steps before accuracy does \citep{shihab2026continual}. Closest to our work, \citet{huang2025beyond} find that iterative self-improvement can make LLMs increasingly overconfident, and compare calibrating once before, once after and at every iteration. Self-training on pseudo-labels has long been known to breed overconfidence through confirmation bias \citep{arazo2020pseudo}. We add a distinction these studies do not draw, between a shift in temperature and a loss of discrimination, and we show that the temperature for each round can be found without labels.

\paragraph{Losses, sampling and recalibration.}
The focal loss \citep{lin2017focal} improves the calibration of classifiers by implicitly regularising their entropy \citep{mukhoti2020focal}. It is not a strictly proper loss, however, so its confidence scores do not equal class probabilities, and a closed-form transformation is needed to recover them \citep{charoenphakdee2021focal}. Our observation that the focal loss leaves a self-trained model underconfident, in a way a single temperature largely corrects, fits this picture. Entropy bonuses penalise confident output distributions directly \citep{pereyra2017regularizing}. Nucleus sampling truncates the low-probability tail of each next-token distribution \citep{holtzman2020curious}, and recent samplers bound the entropy of what remains \citep{baghaei2025toph}; our analysis shows why truncation matters when a model is trained on its own samples. Label-free recalibration of LLMs is an active area \citep{shen2024thermometer,luo2025pretrained}; \citet{luo2025pretrained} fit a fine-tuned model's temperature by aligning its predictions with those of its pre-trained counterpart. We use a simpler relative of that idea, matching the model's mean predictive entropy to that of the calibrated starting model, as a diagnostic and do not claim it as a contribution. Finally, reading confidence from multiple-choice option letters requires removing the model's prior over the letters \citep{zhao2021calibrate,zheng2024mcq}.

\section{Setup}
\label{sec:setup}

\begin{figure*}[t]
\centering
\includegraphics[width=\textwidth]{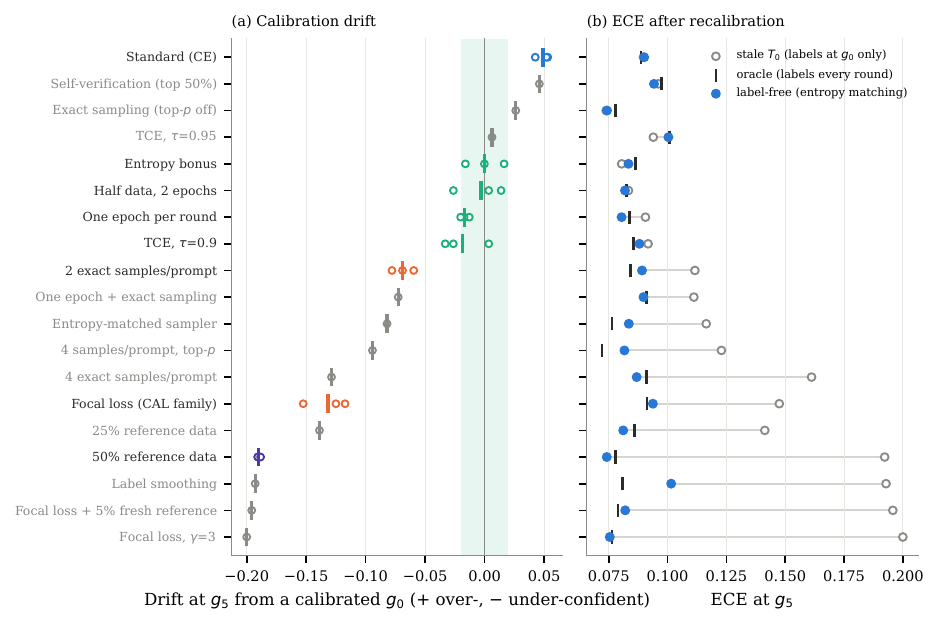}
\caption{All protocols, sorted by calibration drift. (a)~Drift at $\gen{5}$ from a calibrated $\gen{0}$. Circles are seeds, bars are means, and the shaded band marks $|$drift$|\le0.02$. Colours follow Figure~\ref{fig:axes}, and grey labels mark arms with a single seed. (b)~ECE at $\gen{5}$ under three temperatures: one fitted at $\gen{0}$ and never updated (stale), one chosen every round without labels by entropy matching (\S\ref{sec:temp}), and one refitted every round on labels (oracle). The label-free temperature comes within \trackMaxGap\ of the oracle for every protocol except label smoothing.}
\label{fig:protocols}
\end{figure*}

\begin{table*}[t]
\centering\small
\resizebox{\textwidth}{!}{\begin{tabular}{@{}lllllll@{}}
\toprule
 & Entropy kept & Accuracy & AUROC & Calibration drift & \multicolumn{2}{c}{ECE at $g_5$}\\
\cmidrule(l){6-7}
Protocol (3 seeds each) & per round & at $g_5$ & at $g_5$ & from calibrated $g_0$ & stale $T_0$ & label-free\\
\midrule
Standard self-training (CE) & 0.795 {\scriptsize$\pm$0.005} & 0.567 {\scriptsize$\pm$0.004} & 0.704 {\scriptsize$\pm$0.010} & +0.049 {\scriptsize$\pm$0.006} & 0.090 {\scriptsize$\pm$0.006} & 0.090 {\scriptsize$\pm$0.012}\\
\midrule
\multicolumn{7}{@{}l}{\textit{Hold the calibration fixed point}}\\
Half data, 2 epochs & 0.881 {\scriptsize$\pm$0.003} & 0.564 {\scriptsize$\pm$0.010} & 0.721 {\scriptsize$\pm$0.016} & $-$0.003 {\scriptsize$\pm$0.021} & 0.083 {\scriptsize$\pm$0.011} & 0.082 {\scriptsize$\pm$0.007}\\
One epoch per round & 0.944 {\scriptsize$\pm$0.002} & 0.569 {\scriptsize$\pm$0.008} & 0.727 {\scriptsize$\pm$0.020} & $-$0.017 {\scriptsize$\pm$0.004} & 0.091 {\scriptsize$\pm$0.011} & 0.080 {\scriptsize$\pm$0.008}\\
TCE, $\tau{=}0.9$ \citep{shabgahi2025fortifai} & 0.833 {\scriptsize$\pm$0.006} & 0.568 {\scriptsize$\pm$0.005} & 0.714 {\scriptsize$\pm$0.008} & $-$0.019 {\scriptsize$\pm$0.019} & 0.092 {\scriptsize$\pm$0.001} & 0.088 {\scriptsize$\pm$0.009}\\
Entropy bonus \citep{pereyra2017regularizing} & 0.858 {\scriptsize$\pm$0.005} & 0.565 {\scriptsize$\pm$0.008} & 0.715 {\scriptsize$\pm$0.006} & +0.000 {\scriptsize$\pm$0.016} & 0.080 {\scriptsize$\pm$0.005} & 0.083 {\scriptsize$\pm$0.008}\\
\midrule
\multicolumn{7}{@{}l}{\textit{Improve discrimination}}\\
Focal loss (CAL family) \citep{shabgahi2025fortifai} & 0.969 {\scriptsize$\pm$0.005} & 0.565 {\scriptsize$\pm$0.012} & 0.742 {\scriptsize$\pm$0.009} & $-$0.132 {\scriptsize$\pm$0.018} & 0.147 {\scriptsize$\pm$0.009} & 0.094 {\scriptsize$\pm$0.007}\\
Two exact samples per prompt & 0.902 {\scriptsize$\pm$0.005} & 0.563 {\scriptsize$\pm$0.003} & 0.742 {\scriptsize$\pm$0.010} & $-$0.069 {\scriptsize$\pm$0.009} & 0.112 {\scriptsize$\pm$0.009} & 0.089 {\scriptsize$\pm$0.006}\\
\midrule
\multicolumn{7}{@{}l}{\textit{Mix in reference data}}\\
50\% reference data \citep{gerstgrasser2024} & 1.313 {\scriptsize$\pm$0.009} & 0.569 {\scriptsize$\pm$0.009} & 0.720 {\scriptsize$\pm$0.017} & $-$0.190 {\scriptsize$\pm$0.001} & 0.192 {\scriptsize$\pm$0.002} & 0.074 {\scriptsize$\pm$0.007}\\
\bottomrule
\end{tabular}}
\caption{Standard self-training and the seven protocols run with three seeds (mean {\scriptsize$\pm$sd}). \emph{Entropy kept per round} is $\kappa$ from \S\ref{sec:setup}; a value of 1 means no sharpening. \emph{Calibration drift} is mean confidence minus accuracy at $\gen{5}$ under a temperature fitted once at $\gen{0}$ (cross-fitted); positive values are overconfidence. The last two columns give ECE at $\gen{5}$ with that stale temperature and with label-free recalibration. At $\gen{0}$, accuracy is \baseAcc\ and AUROC \baseAUROC. Qwen3-1.7B, all \nItems\ Global-MMLU items.}
\label{tab:protocols}
\end{table*}

\paragraph{The loop.} We start from $M_0$, Qwen3-1.7B \citep{qwen3} with reasoning disabled, and a fixed pool of 3{,}000 questions from TriviaQA \citep{joshi2017triviaqa} and NQ-Open \citep{kwiatkowski2019nq}. In round $g$, the current model $M_g$ answers every question in the pool (temperature 1.0, top-$p$ 0.95, at most 32 tokens). We fine-tune $M_g$ on these 3{,}000 question--answer pairs for two epochs with LoRA \citep{hu2022lora} (rank 16, learning rate $2{\times}10^{-4}$, effective batch size 16) and merge the adapter into the weights to obtain $M_{g+1}$. Each round therefore trains on the previous round's model and on that model's own outputs, the compounding regime in which collapse is usually studied. We evaluate $M_0$ to $M_5$. Each protocol in the paper changes one element of this loop and is run with three seeds unless we say otherwise. The seed changes generation and training, while $M_0$ and its evaluation are shared.

\paragraph{Evaluation.} All models are evaluated on the same \nItems\ English items of Global-MMLU \citep{singh-etal-2025-global}, which do not overlap with the training pool. We read confidence from the multiple-choice answer. After an answer cue, we take the log-probability of each option letter, subtract that letter's mean over the evaluation set to remove the model's positional prior \citep{zhao2021calibrate,zheng2024mcq}, and renormalise. We keep these debiased log-probabilities at full precision, which matters for ranking metrics. Stored as probabilities, the chosen option's confidence is at least 0.999999 for \baseSaturated\ of the \nItems\ items at $\gen{0}$, so many items would tie, and AUROC would reward any method that merely softens the model. With exact log-probabilities, no two items share a confidence value, either at $\gen{0}$ or at $\gen{5}$ of any replicated protocol.

\paragraph{What we measure.} \emph{Accuracy} is the fraction of items whose highest-scoring option is correct. \emph{Discrimination} is the AUROC of the chosen option's log-probability as a predictor of correctness. Because AUROC weighs all pairs of items equally, we also report average precision, which emphasises the items the model is most confident about, and the AUROC of the margin between the two best options (Appendix~\ref{app:disc}). \emph{Calibration drift} measures temperature. $M_0$ answers \baseAcc\ of the items correctly with a mean confidence of \baseConf, so almost all of its raw expected calibration error (ECE) of \baseECEraw\ is overconfidence (\baseOverconf). Any intervention that softens such a model lowers its raw ECE, whether or not its confidence becomes more trustworthy. We therefore calibrate first. We fit a temperature $T_0$ on a random half of the items at $\gen{0}$ (mean $T_0{=}\baseTzero$, held-out ECE \baseECEcal), apply the same $T_0$ at $\gen{5}$, and report mean confidence minus accuracy on the other half; we then swap the halves and average. Positive drift means the model has become more confident than its calibrated start, and negative drift means it has become less confident. When drift is near zero, a temperature fitted once at the start remains valid. Finally, to follow the sharpening during training, we record the model's mean per-token entropy $H_g$ on the answers it generates in round $g$, and summarise it as the entropy kept per round,
\[
\kappa = \Bigl(\frac{H_4}{H_0}\Bigr)^{1/4},
\]
the geometric mean of $H_{g+1}/H_g$ over the four transitions between rounds 0 and 4. A value of 1 means the model does not sharpen.

\paragraph{Protocols.} Besides the standard loop (CE), we vary how the loop trains and samples and which loss it uses. On the loop side, we train for a single pass instead of two epochs, train on a random half of the answers for the same number of optimiser steps (a step-matched control), sample exactly with top-$p$ switched off, and train on several samples per prompt. On the loss side, we use ForTIFAI's truncated cross-entropy at its default threshold $\tau{=}0.9$ and the focal loss from the same confidence-aware family \citep{shabgahi2025fortifai}, an entropy bonus \citep{pereyra2017regularizing}, and label smoothing. We test confidence-based self-verification by keeping the more confident half of each round's answers. We also test the standard defence against collapse by replacing half of each round's answers with the dataset's reference answers \citep{gerstgrasser2024}. The loss and loop variants were first screened with a promotion rule written into the experiment scripts before any of them ran (Appendix~\ref{app:rules}). The protocols we report with three seeds are those the rule promoted, the controls needed to interpret them, and reference mixing, which was part of the original design of the study.

\section{Results}

\subsection{What self-training changes}
\label{sec:temp}

Under the standard loop, accuracy moves from \baseAcc\ to \ceAcc\ over five rounds and AUROC from \baseAUROC\ to \ceAUROC. In each seed, a paired bootstrap over items puts the change in AUROC within noise. What does change is confidence. Measured from a calibrated start, the model ends with a drift of \ceDrift\ ($\pm$\ceDriftsd\ across seeds; Table~\ref{tab:protocols}), so it is clearly more confident than it should be. Raw ECE barely registers this, changing by only \ceRawdECE, because the base model's mean confidence is already \baseConf\ and has little room to rise.

The ranking also changes where it matters most. Average precision, which weights the answers the model is most sure of, falls from \baseAP\ to \ceAP, by between \ceAPdropMin\ and \ceAPdropMax\ in the three seeds, and each seed's 95\% bootstrap interval excludes zero. Among the fifth of the items on which the model is most confident, accuracy falls from \baseTopTwenty\ to \ceTopTwenty. The model becomes surer of everything, and its surest answers contain more mistakes.

A temperature calibrated once at $\gen{0}$ therefore goes stale. Under standard training the cost in ECE is still small at $\gen{5}$ (\ceStale\ with the stale temperature against \ceOracle\ with one refitted on labels), but the cost grows with the drift, and several remedies below drift much further (Figure~\ref{fig:protocols}).

If most of the change is a shift in temperature, one scalar per round should undo it, and it should be possible to find that scalar without labels. We test this with entropy matching, a simple relative of the label-free recalibration of \citet{luo2025pretrained}. In round $g$ we choose the temperature at which the model's mean predictive entropy on the unlabelled evaluation inputs equals that of the calibrated $M_0$. For every protocol except one, the resulting ECE is within \trackMaxGap\ of the ECE obtained by refitting the temperature on labels every round (Figure~\ref{fig:protocols}b; Table~\ref{tab:tracking} in Appendix~\ref{app:tracking}). The exception is label smoothing, run with one seed, where the label-free temperature reaches \lsTracked\ against the oracle's \lsOracle, plausibly because smoothed targets change the shape of the output distribution and not only its temperature. For the focal loss, which drifts far, ECE is \focalStale\ with the stale temperature, \focalTracked\ with the label-free one and \focalOracle\ with the labelled one. The same holds at 8B, where the base model is better calibrated: training on reference data alone takes ECE to \eightrealStale\ under the stale temperature, while the label-free temperature brings it to \eightrealTracked\ (oracle \eightrealOracle).

Self-training thus leaves most of what the model knows and most of its ranking in place, and it moves the model's temperature. The temperature can be recovered without labels, and it is also what most remedies move (\S\ref{sec:remedies}).

\subsection{Why the model sharpens}
\label{sec:mech}
\begin{figure}[t]
\centering
\includegraphics[width=\columnwidth]{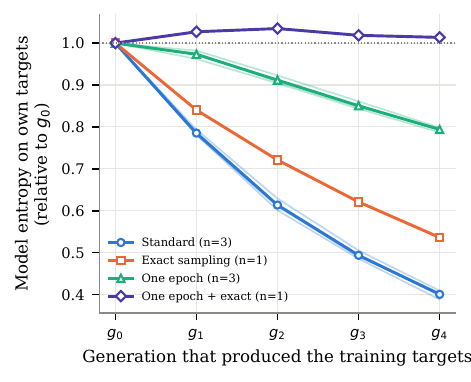}
\caption{The model's mean entropy on the answers it generates for training, relative to round 0. Standard self-training (two epochs, top-$p$ 0.95) sharpens the model steadily. Exact sampling (top-$p$ off) slows the sharpening, a single pass per round slows it more, and the two together keep the entropy flat. Thin lines are individual seeds.}
\label{fig:entropy}
\end{figure}

\paragraph{Self-training is neutral in expectation.} Suppose a model is trained with cross-entropy on exact samples from its own distribution. At the current parameters the expected gradient is zero, because $\mathbb{E}_{y\sim p_\theta}[\nabla_\theta \log p_\theta(y)] = \nabla_\theta \sum_y p_\theta(y) = 0$. The argument concerns the expected update only. Finite samples, the optimiser and the low-rank parameterisation all add noise, so it predicts the absence of a systematic drift rather than the absence of change. Even so, it means that sharpening is not built into self-training, and must come from some way in which the loop departs from this idealisation. The standard loop departs from it in two ways.

First, it \emph{truncates}. With top-$p{=}0.95$, any step whose most likely token already has probability at least 0.95 is sampled greedily. This applies to \probeGreedyGzero\% of steps at $\gen{0}$ and \probeGreedyGfive\% at $\gen{5}$, so the sampler retains \probeKeepGzero\% of the model's entropy at the start and \probeKeepGfive\% after five rounds. Truncation therefore feeds on itself, since a sharper model is sampled more greedily. Second, the loop \emph{repeats}. Each prompt has a single sampled answer, which is fitted for two epochs, so the model is pulled towards that one answer rather than towards its own distribution.

\paragraph{Removing each departure.} Figure~\ref{fig:entropy} follows the model's entropy on its own training answers. Under the standard loop it falls by \ceEntropyLost\% over four rounds, from \ceEntropyFirst\ to \ceEntropyLast\ nats. Sampling exactly, with top-$p$ switched off, raises the entropy kept per round from \ceKept\ to \ancKept, and fitting each answer once instead of twice raises it to \oneepKept. The single-pass effect is not simply a matter of training less, since the step-matched control, which sees half of the answers twice, keeps \halfKept. Repetition therefore accounts for most of the sharpening and truncation for the rest. With both changes in place, a single pass over exact samples, the entropy stays flat at \eoneancEntropyFirst\ nats in round 0 and \eoneancEntropyLast\ in round 4, as the zero-gradient argument predicts. This arm has a single seed, and we read it as a confirmation of the mechanism rather than as a separate result.

\paragraph{Output entropy is not calibration.} Keeping the model's generations at their fixed point does not keep its multiple-choice confidence at its fixed point. The single-pass, exact-sampling arm ends with a drift of \eoneancDrift, and training on several exact samples per prompt overshoots further as the number of samples grows (\ktwoDrift\ for two and \kfourancDrift\ for four; Figure~\ref{fig:protocols}a), even though all of these arms keep more entropy than the standard loop. One reading is consistent with every arm we ran. Two forces act on confidence. Fitting tokens the model already predicts confidently sharpens it, and moving training weight towards tokens it is unsure of softens it, whether the weight moves through disagreeing targets for the same prompt or through down-weighting confident tokens. Single-pass training weakens the first force without adding the second, and it lands closest to the fixed point (\oneepDrift\,$\pm$\,\oneepDriftsd).

\subsection{What the remedies do}
\label{sec:remedies}

Figure~\ref{fig:axes} places each replicated protocol on the two properties, and Table~\ref{tab:protocols} gives the numbers. The protocols fall into two groups, and reference mixing behaves like the second.

\paragraph{Remedies that hold temperature.} TCE at ForTIFAI's default threshold, the entropy bonus and single-pass training all keep a calibrated model calibrated through five rounds, with drifts of \tceDrift, \entDrift\ and \oneepDrift\ against \ceDrift\ for standard training (each $p\le\holdMaxpDrift$). None of them costs accuracy, and none raises AUROC significantly (\tceAUROC, \entAUROC\ and \oneepAUROC\ against \ceAUROC; each $p\ge\holdMinpAUROC$). Some of them do protect the top of the ranking. Single-pass training keeps average precision at \oneepAP, close to its starting value of \baseAP\ and above the \ceAP\ of standard training ($p{=}\oneeppAP$), and TCE keeps part of it (\tceAP, $p{=}\tcepAP$).

\paragraph{Remedies that improve discrimination.} The focal loss from ForTIFAI's family, which down-weights each token by $(1-p_t)^\gamma$, raises AUROC to \focalAUROC\ ($p{=}\focalpAUROC$ against CE), and training on two exact samples per prompt raises it to \ktwoAUROC\ ($p{=}\ktwopAUROC$), again without costing accuracy. Both gains survive the other measures of discrimination. The AUROC of the top-two margin rises by \focaldMargin\ and \ktwodMargin\ ($p{=}\focalpMargin$ and \ktwopMargin), and average precision rises above its starting value, to \focalAP\ and \ktwoAP\ (Appendix~\ref{app:disc}). Both remedies, however, overshoot the fixed point into underconfidence, with drifts of \focalDrift\ and \ktwoDrift. Across protocols, AUROC rises as the model fits its own single answers less (CE \ceAUROC, half data \halfAUROC, single pass \oneepAUROC, focal loss \focalAUROC) or fits several different answers per prompt (\ktwoAUROC), but not when only its most confident tokens are dropped (TCE, \tceAUROC). With three seeds we cannot separate the focal loss from simply training less, since it exceeds the step-matched half-data control by \focalVsHalfdAUROC\ ($p{=}\focalVsHalfp$).

\paragraph{Reference data moves temperature too.} Replacing half of each round's answers with reference answers is the defence that the collapse literature relies on most \citep{gerstgrasser2024,bertrand2024stability}. Judged by raw calibration error it is a clear improvement, changing ECE by \mixRawdECE\ over five rounds. Measured from a calibrated start, however, it ends with a drift of \mixDrift\ ($\pm$\mixDriftsd\ across seeds), deep in underconfidence, with accuracy \mixAcc\ and AUROC \mixAUROC\ ($p{=}\mixpAUROC$ against CE). It does keep average precision near its starting value (\mixAP). Its raw-ECE trajectory, a large improvement after the first round followed by a partial rise, is a large softening followed by partial re-sharpening as the self-generated half of the data is refitted. Like the confidence-aware losses, reference mixing mainly moves temperature, and label-free recalibration brings its ECE to \mixTracked.

\paragraph{Self-verification accelerates overconfidence.} Keeping the more confident half of each round's answers is a simple form of the confidence-based filtering that provably prevents collapse in the setting of \citet{fu2025selfverification}, who score several candidate answers per prompt with the fixed base model. Compared with its matched control, a random half trained with the same budget and the same number of steps, it pushes the model further past its calibrated start (\svtopDrift\ against \halfDrift) and sharpens it faster (\svtopKept\ against \halfKept\ entropy kept per round). This is one seed and one selection rule, so we report it as a warning rather than as a result. Its direction is what the mechanism predicts, since selecting by confidence hands the model its own lowest-entropy answers to fit. Collapse and overconfidence are different targets, and a remedy for one can worsen the other.

\subsection{Getting both}
\label{sec:both}

Because the focal loss overshoots mainly in temperature, label-free recalibration can remove the overshoot. Its ECE falls from \focalStale\ with the stale temperature to \focalTracked\ with the label-free one (Table~\ref{tab:protocols}), the level that the standard loop reaches (\ceTracked). Its advantage in discrimination survives the rescaling. A temperature need not preserve how confidences rank across items, so we recomputed AUROC after recalibration and found \focalAUROCtracked\ for the focal loss against \ceAUROCtracked\ for standard training (\focaldAUROCtracked, $p{=}\focalpAUROCtracked$). A loss that improves discrimination, followed by label-free recalibration, therefore achieves what no single remedy in our study does.

\section{Discussion}

Three lessons follow for how self-training remedies are evaluated and used. The first concerns measurement. Confidence has at least two properties that move independently under self-training, and temperature should be measured from a calibrated start. Raw calibration error rewards any softening of an overconfident model, so by that measure every remedy we tested would look like an improvement, including those that leave the model badly underconfident.

The second concerns the loop itself. The cheapest protection against overconfidence is to fit each self-generated answer once and to sample it without truncation. Both follow from the zero-gradient argument, and neither adds cost. Entropy-aware truncation such as top-H \citep{baghaei2025toph} is a natural alternative to switching truncation off, although our own entropy-matched sampler overshot into underconfidence in a single-seed test (Appendix~\ref{app:all}).

The third concerns remedies that weight or filter data by confidence. They move the two properties in opposite directions, so they should be paired with recalibration rather than trusted to calibrate the model. This agrees with \citet{huang2025beyond}, who find that recalibration largely removes the calibration error that self-improvement introduces, and with the theory of the focal loss, whose confidence scores are known to need a correction \citep{charoenphakdee2021focal}. Our results add that the correction can be a single temperature per round, chosen without labels.

\section{Conclusion}

Recursive self-training leaves a model's accuracy and most of its ranking of answers in place and changes its temperature, while its most confident answers become less reliable. Intervening on the loop traces the sharpening to two design choices, fitting each answer twice and sampling with truncation, and the sharpening disappears when both are removed. The remedies we tested, reference mixing included, either hold temperature or improve discrimination, and none does both. A loss that improves discrimination, followed by label-free recalibration, does.

\section*{Limitations}
Our comparison of remedies uses one model (Qwen3-1.7B), one evaluation set of \nItems\ multiple-choice items, and confidence read from option letters. We repeated the temperature diagnosis at 8B but not the remedies, and scoring the option texts rather than the letters, or measuring confidence in open-ended answers, may behave differently. The loop is pure self-training with no external learning signal, so it isolates what self-training does to confidence but cannot show how remedies interact with a real improvement signal such as a correctness filter. The effects are small in absolute terms (AUROC differences of 0.02--0.04 and drifts of 0.02--0.13), the main comparisons rest on three seeds, and the $p$-values are not corrected across the comparisons we report. Several mechanism arms have a single seed: the flat-entropy arm, the four-sample arms, the entropy-matched sampler, and the self-verification test, which also uses a single selection rule and keep rate. With three seeds we cannot separate the focal loss from training less. We tested one label-free recalibration, entropy matching adapted from \citet{luo2025pretrained}, which we use as a diagnostic; others, such as their divergence-based method or Thermometer \citep{shen2024thermometer}, may track the oracle differently. We did not test entropy-aware truncation inside the loop. The study began with a different hypothesis, that calibration degrades before accuracy, and its planned test is reported in Appendix~\ref{app:rules}. The two-property framing emerged from the data, and the screen of remedies followed a rule written into the experiment scripts before it ran. An earlier Bangla extension was invalidated by a tokenizer truncation default that removed the question from 75\% of passage-bearing prompts. It is excluded, and its diagnosis is documented with the released code.

\section*{Ethics Statement}
All data are public benchmarks used within their licences, and no human subjects were involved. The main risk we see is over-trusting label-free recalibration, which we tested on one model family and on multiple-choice confidence only (see Limitations). The protocols reported here used about 40 GPU-hours on one 16\,GB consumer GPU, and the 8B diagnosis used about 10 hours on a rented 24\,GB GPU.

\paragraph{Artifacts and licences.} The training question pool comes from TriviaQA \citep{joshi2017triviaqa} (\texttt{rc.nocontext}, Apache 2.0) and NQ-Open \citep{kwiatkowski2019nq} (CC BY-SA 3.0), and evaluation uses the English items of Global-MMLU \citep{singh-etal-2025-global} (Apache 2.0). The models are Qwen3-1.7B and Qwen3-8B \citep{qwen3} (Apache 2.0). All were obtained from the Hugging Face Hub, and our research use is consistent with their licences and intended uses. The datasets contain trivia and exam questions; we collect no personal information and add no new content beyond model outputs on these questions.

\paragraph{Reproducibility.} Prompts use the same instruction in every round; it permits but does not request expressions of uncertainty, and reasoning is disabled through the chat template. Training uses a completion-only loss, paged 8-bit AdamW \citep{dettmers2022optimizers}, 3\% warm-up, no weight decay, gradient checkpointing and a 768-token window. Evaluation uses a fixed set of item identifiers shared by every protocol and round. Exact debiased option log-probabilities are stored for every round of the protocols in Table~\ref{tab:protocols}. For earlier standard-loop runs they were recovered at $\gen{0}$ and $\gen{5}$ by re-scoring the saved weights, which reproduced every stored probability exactly. Bootstrap intervals are paired over items, and $p$-values across seeds come from Welch $t$-tests with three seeds per group, uncorrected for multiple comparisons. The code and the experiment scripts, including the decision rules of Appendix~\ref{app:rules}, are available at the repository linked in the introduction; we will release the per-item records with the final version.

\paragraph{Software.} We use PyTorch 2.10, Hugging Face Transformers 5.15, PEFT 0.20 for LoRA, bitsandbytes 0.50 for the 8-bit optimiser and Datasets 5.0; the statistics use NumPy 1.26 and SciPy 1.14.

\bibliography{references}

\appendix

\section{Decision rules and their outcomes}
\label{app:rules}
\paragraph{Original hypothesis.} The study was designed to test whether calibration degrades at an earlier round than accuracy. The test was written into the analysis code before the primary runs, although it was not publicly pre-registered: a paired item-level bootstrap per round, a minimum effect of 0.02, Holm correction across calibration metrics, and an accuracy floor relative to chance. On the standard loop the hypothesis was supported in one of three seeds and undetermined in the other two, because the raw calibration drift was below the minimum effect. Accuracy never degraded in any run.
\paragraph{Protocol screen.} Because temperature can be corrected after training, the screen's primary endpoint was discrimination, and the rule was written into the experiment scripts before any screened protocol ran. A protocol was promoted if its $\gen{5}$ AUROC on exact log-probabilities exceeded that of the standard loop by at least 0.03, with accuracy no more than 0.01 below it. The focal loss passed at one seed and replicated on the three-seed mean. Single-pass training passed at one seed but did not replicate on the mean (\oneepAUROC); we report it for its effect on calibration, which did replicate. The step-matched half-data control, TCE at $\tau{=}0.9$, the entropy bonus and two samples per prompt were added as controls and replications.

\section{All protocols}
\label{app:all}
\begin{table*}[h]
\centering\small
\resizebox{\textwidth}{!}{\begin{tabular}{@{}lcccccc@{}}
\toprule
Protocol & Seeds & Entropy kept/round & Acc.\ $g_5$ & AUROC $g_5$ & Calibration drift & Raw $\Delta$ECE\\
\midrule
Standard self-training (CE) & 3 & 0.795 & 0.567 & 0.704 & +0.049 & +0.015\\
Half data, 2 epochs & 3 & 0.881 & 0.564 & 0.721 & $-$0.003 & +0.007\\
One epoch per round & 3 & 0.944 & 0.569 & 0.727 & $-$0.017 & $-$0.001\\
TCE, $\tau{=}0.9$ \citep{shabgahi2025fortifai} & 3 & 0.833 & 0.568 & 0.714 & $-$0.019 & +0.005\\
Entropy bonus \citep{pereyra2017regularizing} & 3 & 0.858 & 0.565 & 0.715 & +0.000 & +0.014\\
Focal loss (CAL family) \citep{shabgahi2025fortifai} & 3 & 0.969 & 0.565 & 0.742 & $-$0.132 & $-$0.034\\
Two exact samples per prompt & 3 & 0.902 & 0.563 & 0.742 & $-$0.069 & $-$0.010\\
50\% reference data \citep{gerstgrasser2024} & 3 & 1.313 & 0.569 & 0.720 & $-$0.190 & $-$0.072\\
\midrule
Exact sampling (top-$p$ off) & 1 & 0.856 & 0.572 & 0.705 & +0.026 & +0.006\\
One epoch + exact sampling & 1 & 1.003 & 0.582 & 0.703 & $-$0.073 & $-$0.031\\
Four exact samples per prompt & 1 & 0.919 & 0.573 & 0.737 & $-$0.129 & $-$0.031\\
Four samples per prompt, top-$p$ & 1 & 0.856 & 0.560 & 0.746 & $-$0.094 & $-$0.015\\
Entropy-matched sampler & 1 & 0.921 & 0.568 & 0.706 & $-$0.082 & $-$0.003\\
TCE, $\tau{=}0.95$ & 1 & 0.797 & 0.567 & 0.716 & +0.006 & +0.010\\
Focal loss, $\gamma{=}3$ & 1 & 1.250 & 0.552 & 0.743 & $-$0.200 & $-$0.117\\
Focal loss + 5\% fresh reference & 1 & 1.118 & 0.562 & 0.731 & $-$0.196 & $-$0.080\\
Label smoothing & 1 & 1.596 & 0.560 & 0.699 & $-$0.193 & $-$0.040\\
25\% reference data & 1 & 1.162 & 0.553 & 0.730 & $-$0.139 & $-$0.019\\
Self-verification (top 50\%) & 1 & 0.830 & 0.562 & 0.725 & +0.046 & +0.020\\
\bottomrule
\end{tabular}}
\caption{Every protocol we ran, including those with a single seed. Raw $\Delta$ECE is the change in uncalibrated ECE from $\gen{0}$ to $\gen{5}$; it favours any protocol that softens the overconfident base model.}
\label{tab:all}
\end{table*}

\section{Discrimination under other measures}
\label{app:disc}
Table~\ref{tab:disc} repeats the comparison of discrimination with three further measures. Margin AUROC scores each item by the gap between its two highest option log-probabilities instead of by the chosen option's log-probability. Average precision ranks items by the chosen option's log-probability and treats correct answers as positives, so it weights the items the model is most confident about. The last column gives the accuracy among the 20\% of items with the highest confidence. The focal loss and two samples per prompt improve on standard training under every measure. Average precision also separates standard training from some of the protocols that hold temperature, which AUROC does not: standard training loses precision among its most confident answers, while single-pass training, TCE and reference mixing keep more of it.
\begin{table}[h]
\centering\small
\resizebox{\columnwidth}{!}{\begin{tabular}{@{}lllll@{}}
\toprule
Protocol ($g_5$; 3 seeds each) & AUROC & Margin AUROC & Average precision & Acc.\ of top 20\%\\
\midrule
$g_0$ (shared start) & 0.711 & 0.710 & 0.782 & 0.867\\
\midrule
Standard self-training (CE) & 0.704 {\scriptsize$\pm$0.010} & 0.704 {\scriptsize$\pm$0.011} & 0.741 {\scriptsize$\pm$0.011} & 0.800 {\scriptsize$\pm$0.030}\\
Half data, 2 epochs & 0.721 {\scriptsize$\pm$0.016} & 0.720 {\scriptsize$\pm$0.016} & 0.757 {\scriptsize$\pm$0.008} & 0.817 {\scriptsize$\pm$0.008}\\
One epoch per round & 0.727 {\scriptsize$\pm$0.020} & 0.726 {\scriptsize$\pm$0.020} & 0.781 {\scriptsize$\pm$0.011} & 0.844 {\scriptsize$\pm$0.013}\\
TCE, $\tau{=}0.9$ \citep{shabgahi2025fortifai} & 0.714 {\scriptsize$\pm$0.008} & 0.713 {\scriptsize$\pm$0.007} & 0.767 {\scriptsize$\pm$0.004} & 0.831 {\scriptsize$\pm$0.024}\\
Entropy bonus \citep{pereyra2017regularizing} & 0.715 {\scriptsize$\pm$0.006} & 0.714 {\scriptsize$\pm$0.005} & 0.752 {\scriptsize$\pm$0.010} & 0.831 {\scriptsize$\pm$0.005}\\
Focal loss (CAL family) \citep{shabgahi2025fortifai} & 0.742 {\scriptsize$\pm$0.009} & 0.739 {\scriptsize$\pm$0.009} & 0.811 {\scriptsize$\pm$0.004} & 0.908 {\scriptsize$\pm$0.022}\\
Two exact samples per prompt & 0.742 {\scriptsize$\pm$0.010} & 0.741 {\scriptsize$\pm$0.011} & 0.798 {\scriptsize$\pm$0.010} & 0.883 {\scriptsize$\pm$0.025}\\
50\% reference data \citep{gerstgrasser2024} & 0.720 {\scriptsize$\pm$0.017} & 0.716 {\scriptsize$\pm$0.018} & 0.784 {\scriptsize$\pm$0.011} & 0.881 {\scriptsize$\pm$0.039}\\
\bottomrule
\end{tabular}}
\caption{Discrimination at $\gen{5}$ under four measures (mean {\scriptsize$\pm$sd} over three seeds). All are computed from exact debiased log-probabilities, which contain no ties.}
\label{tab:disc}
\end{table}

\section{Label-free recalibration at 1.7B and 8B}
\label{app:tracking}
\begin{table}[h]
\centering\small
\resizebox{\columnwidth}{!}{\begin{tabular}{@{}llccc@{}}
\toprule
Model & Protocol & Stale $T_0$ & Label-free & Oracle\\
\midrule
1.7B & Standard self-training (CE) & 0.090 & 0.090 & 0.089\\
1.7B & Half data, 2 epochs & 0.083 & 0.082 & 0.083\\
1.7B & One epoch per round & 0.091 & 0.080 & 0.084\\
1.7B & TCE, $\tau{=}0.9$ \citep{shabgahi2025fortifai} & 0.092 & 0.088 & 0.085\\
1.7B & Entropy bonus \citep{pereyra2017regularizing} & 0.080 & 0.083 & 0.086\\
1.7B & Focal loss (CAL family) \citep{shabgahi2025fortifai} & 0.147 & 0.094 & 0.091\\
1.7B & Two exact samples per prompt & 0.112 & 0.089 & 0.084\\
1.7B & 50\% reference data \citep{gerstgrasser2024} & 0.192 & 0.074 & 0.078\\
\midrule
8B & Standard (CE), seed 1 & 0.086 & 0.080 & 0.084\\
8B & Standard (CE), seed 2 & 0.088 & 0.073 & 0.070\\
8B & Reference data only, seed 1 & 0.247 & 0.057 & 0.059\\
8B & 50\% reference data, seed 1 & 0.225 & 0.075 & 0.075\\
\bottomrule
\end{tabular}}
\caption{ECE at $\gen{5}$ with a temperature fitted once at $\gen{0}$ (stale), chosen without labels by entropy matching (\S\ref{sec:temp}), and refitted on labels every round (oracle), with two-fold cross-fitting. At 8B the probabilities were not stored at full precision, so the label-free and oracle temperatures there use the same saturated scores.}
\label{tab:tracking}
\end{table}

\end{document}